\documentclass[letterpaper]{article} 
\usepackage{aaai25}  
\usepackage{times}  
\usepackage{helvet}  
\usepackage{courier}  
\usepackage[hyphens]{url}  
\usepackage{graphicx} 
\usepackage{natbib}  
\usepackage{caption} 
\usepackage{algorithm}
\usepackage{algorithmic}

\usepackage{newfloat}
\usepackage{listings}
\DeclareCaptionStyle{ruled}{labelfont=normalfont,labelsep=colon,strut=off} 
\floatstyle{ruled}
\newfloat{listing}{tb}{lst}{}
\floatname{listing}{Listing}
\usepackage{amssymb}            
\usepackage{amsmath}
\usepackage{amsthm}
\usepackage{color}
\usepackage{mathtools}          
\usepackage{mathrsfs}           
\usepackage{subcaption}         
\usepackage{url}      
\usepackage{comment}
\newtheorem{definition}{Definition}
\newtheorem{lemma}{Lemma}

\newtheorem{theorem}{Theorem}
\newtheorem{proposition}{Proposition}
\newtheorem{fact}{Fact}

\usepackage{scalerel}

\title{On Shallow Planning Under Partial Observability}
\author{
    Randy Lefebvre\textsuperscript{\rm 1},
    Audrey Durand\textsuperscript{\rm 1,\rm 2}
}
\affiliations{
    \textsuperscript{\rm 1}Université Laval\\
    \textsuperscript{\rm 2}Canada CIFAR AI Chair\\
    randy.lefebvre.1@ulaval.ca, audrey.durand@ift.ulaval.ca

}

\newcommand{\valueState}[2]{V_{#1}^{#2}}
\newcommand{\transitionMatrix}[2]{[P_{#1}^{#2}]}
\newcommand{\ets}{e_{s}^{\top}}
\newcommand\wh[1]{\hstretch{2}{\hat{\hstretch{.5}{#1}}}}

\begin{document}

\maketitle

\begin{abstract}
Formulating a real-world problem under the Reinforcement Learning framework involves non-trivial design choices, such as selecting a discount factor for the learning objective (discounted cumulative rewards), which articulates the planning horizon of the agent. This work investigates the impact of the discount factor on the bias-variance trade-off given structural parameters of the underlying Markov Decision Process. Our results support the idea that a shorter planning horizon might be beneficial, especially under partial observability.
\end{abstract}

\section{Introduction}

Reinforcement Learning (RL) has had tremendous success on Atari games~\citep{mnih_playing_2013}, yet applications of RL in the real-world remain limited~\citep{dulac-arnold_challenges_2021}. This complexity is due to many challenges such as sample efficiency of RL methods~\citep{yu2018towards}, risk/safety issues~\citep{Gu_Yang_Du_Chen_Walter_Wang_Yang_Knoll_2023} and partial observability~\citep{sondik1978optimal,francois-lavet_overfitting_2019,kaelbling1998planning}. Formulating a real-world problem under the RL framework also involves several non-trivial decisions such as selecting a state/action space (especially when these are continuous), formulating a reward function, and formulating the learning objective~\citep{hare2019dealing,devidze2021explicable}. The learning objective usually corresponds to the discounted cumulative rewards, which depends on a discount factor articulating the considered planning horizon when attributing values to states and actions~\citep{sutton_reinforcement_2018}. This objective is useful since it can reduce the search space intuitively by giving less credit to future rewards and actions. In toy and simulated environments, early practitioners tend to use large discount factor values often found in the RL literature on Atari~\citep{Kaiser_Babaeizadeh_Milos_Osinski_Campbell_Czechowski_Erhan_Finn_Kozakowski_Levine_etal._2024, mnih_playing_2013}. This equates to considering very long planning horizons. On the other hand, real-world applications tend to formulate sequential decision-making problems under the contextual bandit setting (i.e. with a myopic agent w.r.t. the planning horizon) in response to low data regimes~\citep{bastani2020online,Ding,durand2018contextual}.

The impact of reducing the planning horizon has been studied previously, and bounds on the resulting bias-variance trade-off on the state value functions have been proposed~\citep{amit_discount_2020,jiang2015dependence}. Unfortunately, these results provide loose bounds that do not consider the structure of the underlying Markov Decision Process (MDP) and thus fail to capture its impact on the \textit{optimal planning horizon}. The optimal planning horizon can be defined using the discount factor $\gamma$ which minimizes the planning loss (see Eq. \ref{eq:planning_loss}), i.e. the one which can extract the best policy possible considering the limited amount of data. Distinct results involving the structure of MDPs~\citep{jiang2016structural, gheshlaghi_azar_minimax_2013,wu2023uniform,he2021uniform} have been achieved separately, but these insights have never been brought together.

\paragraph{Contributions}
In this work, we introduce new results on the bias-variance trade-off that explicitly depend on high-level structural parameters of the underlying MDP. More importantly, our results touch on Partially Observable MDPs (POMDPs), providing the first insights supporting the advantage of considering short horizons in the learning objective for practical applications under partial observability. We support and illustrate the theory with experiments, hoping that this can open the door to choices in learning objectives that are better adapted to real-world RL applications. Finally, we provide the open-source code\footnote{\url{https://github.com/GRAAL-Research/shallow-planning-partial-observability}} for all our experiments to ensure reproducibility and offer a framework that practitioners can modify to better understand the impact of partial observability on their specific applications.

\section{Fully Observable Setting}
\label{sec:mdp}

An MDP can be described as a tuple $(\mathcal S, \mathcal A, P, R)$, where $\mathcal S$ is a finite state space, $\mathcal A$ is a finite action space, $P$ : $\mathcal S \times \mathcal A \times \mathcal S \mapsto [0,1]$ is a transition function, and $R$ : $\mathcal S \times \mathcal A \mapsto [0,R_{\text{max}}]$ is a reward function, with $R_{\text{max}}$ denoting the maximal reward obtainable in the MDP. On each time step $t\in \mathbb N_0$, the current state $S_t\in\mathcal S$ is observed, an action $A_t\in\mathcal A$ is played, the environment transitions into next state $S_{t+1}$ (using $P$) and generates an observed reward $R_{t+1}$ (using $R$). Given an MDP (tuple) $M$, the value of state $s\in\mathcal S$ under policy $\pi$ : $\mathcal S \mapsto \mathcal A$ is the expected sum of discounted rewards obtained by selecting actions according to policy $\pi$ from state $s$:
\begin{align*}
    V_{M,\gamma}^\pi(s) = \mathbb{E}_\pi\left[ \sum_{k=0}^\infty \gamma^{k} R_{t+k+1} | S_t = s\right],
\end{align*}
where the discount factor $\gamma \in [0, 1]$ controls the planning horizon $1/(1-\gamma)$, i.e. the credit assigned to action $A_t$ for future rewards.
The goal of a learning agent is to find the optimal policy $\pi_{M,\gamma}^*$ that maximizes $V_{M,\gamma}^{\pi} (s)$ for all states $s\in\mathcal S$. We consider a setting where this policy is unique. We use $V_{M,\gamma}^{\pi} \in \mathbb{R}^{|\mathcal S|}$ to denote the vector of state values. A notation table can be found in Appendix A.

\paragraph{Blackwell Discount Factor} Practitioners often believe using a higher discount factor will result in a better policy. While this is true with an infinite amount of data, it is rarely the case when using RL in practice. It has even been shown previously that there always exists a discount factor $\gamma_{Bw}$ such that for any $\gamma \geq \gamma_{Bw}$, we have $V_{M,\gamma}^{\pi_{M,\gamma}} =  V_{M,\gamma}^{\pi_{M,\gamma_{Bw}}}$ when $|\mathcal{S}| < \infty$ and $|\mathcal{A}| < \infty$ (\citet{gr-Clément_Petrik_2023}, Thm.~3.2). We refer to $\gamma_{Bw}$ as the \textit{Blackwell discount factor} and we say that optimal policies with discount factor $\gamma \geq \gamma_{Bw}$ are optimal according to the Blackwell optimality criterion. This concept closely resembles the idea of \textit{effective planning horizon} by \citet{laidlaw2023bridging}, with the effective horizon defined as $1/(1-\gamma_{Bw})$ instead of a number of look ahead steps.

\paragraph{Planning Loss} The planning loss captures the impact of using 
$\gamma < \gamma_{Bw}$ given that the planning is performed in an approximate model of the environment\footnote{We remain agnostic to how $\wh{M}$ is 
computed and how the data is collected as literature on the topic is abundant~\citep{gheshlaghi_azar_minimax_2013,wu2023uniform,he2021uniform}.} $\wh{M}$ with $\wh{M} \approx M$: 
\begin{align}
    &\lVert V_{M,\gamma_{Bw}}^{\pi_{M,\gamma_{Bw}}^{*}} -V_{M,\gamma_{Bw}}^{\pi_{\wh{M},\gamma}^{*}} \rVert_{\infty} \\
    &= \lVert V_{M,\gamma_{Bw}}^{\pi_{M,\gamma_{Bw}}^{*}} - V_{M,\gamma_{Bw}}^{\pi_{M,\gamma}^{*}} + V_{M,\gamma_{Bw}}^{\pi_{M,\gamma}^{*}} - V_{M,\gamma_{Bw}}^{\pi_{\wh{M},\gamma}^{*}} \rVert_{\infty} \nonumber\\ \label{eq:planning_loss}
    &\leq  \underbrace{ \lvert\lvert V_{M,\gamma_{Bw}}^{\pi_{M,\gamma_{Bw}}^{*}} - V_{M,\gamma_{Bw}}^{\pi_{M, \gamma}^{*}} \rvert\rvert_{\infty} }_{\text{bias}}
    + \underbrace{ \lvert\lvert V_{M,\gamma_{\text{Bw}}}^{\pi_{M,\gamma}^{*}} - V_{M,\gamma_{Bw}}^{\pi_{\wh{M}, \gamma}^{*}} \rvert\rvert_{\infty}}_{\text{variance}}.
\end{align}
This decomposition offers insight into two components which can affect the quality of the policy obtained when planning on an approximate model of the environment using a shallow planning horizon ($\gamma < \gamma_{Bw}$).  The \textit{bias} denotes the loss in value function (evaluated on the true MDP $M$ and with the Blackwell planning horizon) when using a policy that is optimal with a shallow planning horizon instead of using a policy that is optimal with $\gamma_{Bw}$. On the other hand, the \textit{variance} captures the impact of optimizing the policy under an approximate model $\wh{M}$ with a shallow planning horizon and will tend to $0$ with more data. This decomposition is different from previous work~\citep{jiang2015dependence} and has the advantage of being interpretable as a bias-variance trade-off from the PAC-learning literature. We can compare the bias to the approximation error and the variance to the estimation error~\citep{shalev-shwartz_understanding_2014}.

Structural parameters have been introduced previously to characterize the difficulty of an MDP.
\begin{definition}[Value-function variation~\citep{jiang2016structural}]
\label{def:value_function_variation}
Given an MDP $M$ and discount factor $\gamma$, the value function variation captures the largest difference between the values of two different states when following the optimal policy:
        $$\kappa_{M,\gamma}=\max_{s,s^{\prime}\in \mathcal S}\left|V_{M,\gamma}^{\pi_{M,\gamma}^*}(s)-V_{M,\gamma}^{\pi_{M,\gamma}^*}(s^{\prime})\right| \leq \frac{R_{max}}{1-\gamma}.
    $$
    Note that the value-function is evaluated with the same discount factor as used by the policy.
\end{definition}
The value-function variation provides insight on the impact of starting in certain states over others in the MDP. A low value indicates that the starting state is not important to consider for predicting future rewards \citep{jiang2016structural}. 
%
\begin{definition}[Action variation~\citep{jiang2016structural}]
\label{def:action_variation}
Given an MDP $M$ with transition probabilities $P$, the action variation captures how much actions can impact state transitions:
     $$\delta_M=\max_{s\in \mathcal S}\max_{a,a'\in \mathcal A}\left\|P(\cdot|s,a)-P(\cdot|s,a')\right\|_1.
    $$
\end{definition}
%
If the action variation is equal to $0$, the agent cannot influence the state transitions (and therefore future rewards). In this case, we would expect the problem to be safely (and efficiently) formulated under the contextual bandit setting, which corresponds to using a myopic agent ($\gamma =0$). The maximal value for this $L_1$ distance is $2$, which often happens under deterministic settings.
This is validated by our numerical experiments below.

Using Definitions~\ref{def:value_function_variation} and~\ref{def:action_variation},  the following bound on the bias was introduced~\citep{jiang2016structural}:

\begingroup
\fontsize{9}{8}
\begin{align}
    \label{eq:bias_bound_jiang2016}
    \lvert \lvert \underbrace{V_{M,\gamma_{Bw}}^{\pi_{M,\gamma_{Bw}}^{*}} - V_{M,\gamma_{Bw}}^{\pi_{M, \gamma}^{*}}}_{\text{bias}} \rvert \rvert_{\infty} \leq \frac{\delta_{M}/2\cdot\kappa_{M,\gamma}(\gamma_{Bw}-\gamma)}{(1-\gamma_{Bw})(1-\gamma_{Bw}(1-\delta_{M}/2))}.
\end{align}
\endgroup
Unfortunately, the action variation is not sensitive to the planning horizon of the agent compared with the Blackwell planning horizon. Moreover, unlike the bias, there is no current bounds for the variance. We address these limitations with Lem.~\ref{lemma:variance} and Eq.~\ref{eq:bias_bound_jiang2016}, which will allow us to obtain a new bound on the planning loss~(upcoming Thm.~\ref{thm:planning_loss}).

\subsection{Improving the Bias Bound}
\label{sec:mdp:bias}

In order to consider the planning horizon in the existing bias bound~(Eq.~\ref{eq:bias_bound_jiang2016}), we introduce the following definitions:

\begin{definition}[Discordant state-action pairs] The set of state-action pairs in an MDP $M$ where policies $\pi$ and $\pi'$ differ:
\label{def:discordantSApair}
    \begin{equation*}
        \mathcal{Z}_{M}(\pi \neq \pi') = \{(s,a) \in \mathcal S \times \mathcal A: \pi (s) \neq \pi'(s) , \ \pi'(s) = a \}.
    \end{equation*}
\end{definition}
This new set will be used to capture the impact of a shallow-horizon policy on the action variation:
\begin{definition}[Horizon-sensitive action variation]
\label{def:horizonSensitiveActionVariation}
The most important difference in transition probabilities induced by using discount factor $\gamma$ instead of $\gamma_{Bw}$ on an MDP $M$:
\begingroup
\fontsize{9}{8}
    \begin{equation*}
        \delta_{M,\gamma} = \max_{(s,a) \in \mathcal Z_M(\pi^*_{M,\gamma} \neq \pi^*_{M,\gamma_{Bw}} )} \lVert P(\cdot|s,\pi_{M,\gamma}^*(s))-P(\cdot|s,a)\rVert_1.
    \end{equation*}
\end{definition}
\endgroup

\label{app:BiasExtension}

The implementation of the action variations in proofs is to bound the difference in transition probabilities between two different policies. The highest possible bound is given by prior results (Def.~\ref{def:action_variation}), but we tighten this result by simply considering states for which the policies are unequal instead of all states. This has the benefit of being $0$ when the policy is evaluated with a discount factor above the Blackwell. By building on previous analysis~\citep{jiang2016structural}, we can obtain the following result to characterize the impact of optimizing the policy with a shallow horizon on $k$-steps transition probabilities.
\begin{proposition}[Horizon-sensitive transition probabilities distance]
\label{prop:l1dist:horizon_sensitive}
Given an MDP $M$, let $P_{s, k}^\pi$ denote the vector of the transition probabilities from state $s\in\mathcal S$ to every possible states when following policy $\pi$ for $k\geq 1$ time steps. The transition probabilities when following the policy that is optimal for a shallow planning horizon ($\gamma < \gamma_{Bw}$) instead of following the policy that is optimal for the Blackwell planning horizon is bounded by:
\begin{equation*}
    \begin{aligned}
        \lVert P_{s,k}^{\pi^*_{M,\gamma}} - P_{s,k}^{\pi^*_{M,\gamma_{Bw}}} \rVert_1 \leq 2 - 2(1- \delta_{M,\gamma}/2)^k.
    \end{aligned}
\end{equation*}
\end{proposition}

Prop.~\ref{prop:l1dist:horizon_sensitive} can be used to extend Eq.~\ref{eq:bias_bound_jiang2016} using the horizon-sensitive structural parameters:
\begin{align}
    \label{eq:bias-extension}
    & \lvert \lvert \underbrace{V_{M,\gamma_{Bw}}^{\pi_{M,\gamma_{Bw}}^{*}} - V_{M,\gamma_{Bw}}^{\pi_{M, \gamma}^{*}}}_{\text{bias}} \rvert \rvert_{\infty} \leq \nonumber\\
    & \qquad \qquad \qquad \frac{\delta_{M,\gamma}/2\cdot\kappa_{M,\gamma}(\gamma_{Bw}-\gamma)}{(1-\gamma_{Bw})(1-\gamma_{Bw}(1-\delta_{M,\gamma}/2))},
\end{align}
where we use the term $\delta_{M,\gamma}$ (instead of $\delta_M$) to capture the divergence between the shallow and optimal-horizon policies. Since the set offered by Def.~\ref{def:discordantSApair} is smaller than the set of all state-action pairs, Eq.~\ref{eq:bias-extension} is tighter than Eq.~\ref{eq:bias_bound_jiang2016}. See Appendix B for the complete proofs.

\subsection{Controlling the Variance}
\label{sec:mdp:variance}

We now introduce new definitions and results to bound the variance (in Eq.~\ref{eq:planning_loss}). Recall that the variance captures the impact of learning an optimal policy on an (empirical) approximation $\wh{M}$ of a true MDP $M$ when using a shallow planning horizon ($\gamma < \gamma_{Bw}$). To this end, it is convenient to isolate the variance that does not depend on the shallow planning.

\begin{definition}[Variance due to model approximation] 
\label{def:variancemodelApprox}
The maximum difference in value-function due to the approximate model $\wh{M}$:
    \begin{equation*}
        \hat{\epsilon} = \lVert V_{M,\gamma}^{\pi_{M,\gamma}^*} - V_{M,\gamma}^{\pi_{\wh{M},\gamma}^*} \rVert_{\infty}.
    \end{equation*}
\end{definition}
%
%
This term can be upper-bounded into the following results by using known settings in the PAC literature~\citep{gheshlaghi_azar_minimax_2013,wu2023uniform,he2021uniform}. We can also use the discordant state-action pairs (Def.~\ref{def:discordantSApair}) to capture the action variation resulting from having optimized the policy on an approximation $\wh{M}$ of a true MDP $M$.
\begin{definition}[Empirical action variation]
\label{def:empirical_action_variation}
The most important difference in transition probabilities when following the policy optimal on an MDP $M$ vs the policy optimal on an approximate model $\wh{M}$:
\begingroup
\fontsize{9}{8}
    \begin{equation*}
         \hat{\delta}_{M,\gamma} = \max_{(s,a) \in \mathcal{Z}_{M}(\pi^*_{M,\gamma} \neq \pi^*_{\wh{M},\gamma} )} \left\|P(\cdot|s,\pi_{M,\gamma}^*(s))-P(\cdot|s,a)\right\|_1.
    \end{equation*}
\endgroup
\end{definition}

The improvement over the action variation (Def.~\ref{def:action_variation}) is that it will tend towards $0$ as $\wh{M} \approx M$ which is desirable in a bound on the variance. As was done previously in Definition~\ref{def:horizonSensitiveActionVariation}, we can also build on previous analysis~\citep{jiang2016structural} to obtain the following result to characterize the impact of optimizing the policy with an approximate model $\wh{M}$ on $k$-steps transition probabilities.
\textbf{\begin{proposition}[Empirical transition probabilities distance]
\label{prop:l1dist:empirical}
Given an MDP $M$ and an approximate model $\wh{M}$. Let $P_{s, k}^\pi$ denote the vector of the transition probabilities from state $s\in\mathcal S$ to every possible states when following policy $\pi$ for $k\geq 1$ time steps. For a given planning horizon, the transition probabilities when following the optimal policy on $\wh{M}$ instead of following the optimal policy on $M$ is bounded by:
\begin{equation*}
    \begin{aligned}
        \lVert P_{s,k}^{\pi^*_{M,\gamma}} - P_{s,k}^{\pi^*_{\wh{M},\gamma}} \rVert_1 \leq 2 - 2(1-\hat \delta_{M,\gamma}/2)^k.
    \end{aligned}
\end{equation*}
\end{proposition}}
%
%
Prop.~\ref{prop:l1dist:empirical} can be used with Def.~\ref{def:discordantSApair} and~\ref{def:empirical_action_variation} to obtain the following bound on the variance (see Appendix C).
\begin{lemma}[Variance]
\label{lemma:variance}
Consider optimal policies computed with a shallow planning horizon ($\gamma < \gamma_{Bw}$) on an MDP $M$ and an approximate model $\wh{M}$. The difference between their value-functions evaluated on $M$ with discount factor $\gamma_{Bw}$ is bounded by:
\begin{align*}
     &\lVert V_{M,\gamma_{Bw}}^{\pi_{M,\gamma}^{*}} - V_{M,\gamma_{Bw}}^{\pi_{\wh{M}, \gamma}^{*}} \rVert_{\infty} \\
     &\leq \hat \epsilon\left(\frac{1-\gamma}{1-\gamma_{Bw}}\right) + \frac{\hat{\delta}_{M,\gamma}/2\cdot\kappa_{M,\gamma}(\gamma_{Bw}-\gamma)}{(1-\gamma_{Bw})(1-\gamma_{Bw}(1-\hat{\delta}_{M,\gamma}/2))}.
\end{align*}
\end{lemma}

This bound is interesting because it becomes tighter when the empirical action variation $\hat{\delta}_{M,\gamma}$ or the value function variation $\kappa_{M,\gamma}$ decrease. We can then deduce that a problem with a low value in these structural parameters lowers both the bias (Eq.~\ref{eq:bias-extension}) and the variance. Finally, the use of the empirical action variation (Def.~\ref{def:empirical_action_variation}) gives rise to a bound that is coherent in convergence, as it will tend towards $0$ as the amount of data increases.

%
%
%
%

\subsection{A New Bound on the Planning Loss}
\label{sec:mdp:planning_loss}
%
By combining the extended bias bound (Eq.~\ref{eq:bias-extension}) with our novel bound on the variance (Lem.~\ref{lemma:variance}), we obtain the following bound on the planning loss (Eq.~\ref{eq:planning_loss}). See Appendix D for the complete proof.
%

%
%
\begin{theorem}[Planning loss]
\label{thm:planning_loss}
Given an MDP $M$, its Blackwell discount factor $\gamma_{Bw}$, and an approximate model $\wh{M}$. The planning loss is bounded by:
    \begin{align*}
&\lVert V_{M,\gamma_{Bw}}^{\pi_{M,\gamma_{Bw}}^{*}}-V_{M,\gamma_{Bw}}^{\pi_{\wh{M},\gamma}^{*}}\rVert_{\infty} \\
&\leq \kappa_{M,\gamma}\left(\frac{\gamma_{Bw} - \gamma}{1-\gamma_{Bw}}\right)\left( \frac{\delta_{M,\gamma}/2}{1-\gamma_{Bw}(1-\delta_{M,\gamma}/2)} \right) \\ &\quad + \kappa_{M,\gamma}\left(\frac{\gamma_{Bw} - \gamma}{1-\gamma_{Bw}}\right) \left(\frac{\hat{\delta}_{M,\gamma}/2}{1-\gamma_{Bw}(1-\hat{\delta}_{M,\gamma}/2)} \right) \\
&\quad + \hat{\epsilon}\left(\frac{1-\gamma}{1-\gamma_{Bw}}\right).
\end{align*}
\end{theorem}

This result provides insight into how structural parameters affect not only the bias, but also the variance. For instance, a problem with action variation $\delta_M \approx 0$ has low variance due to the limited impact of the policy over the state value (agent actions do not impact transition probabilities). Similarly to prior work \citep{jiang2015dependence}, although the applicability of this result is limited by not having access to the true model $M$, it remains a helpful guide to design heuristics and better understand how one could decide a discount factor. For example, it justifies framing recommender systems as contextual bandits when the outcome of future recommendations do not depend on current recommendations, which translates into a low value-function variation $\kappa_{M,\gamma}$.
Thm.~\ref{thm:planning_loss} is tighter than the current existing bound~\citep{jiang2015dependence} under the following condition on the quality of the model approximation $\wh{M}$ (see Appendix E): 
\begin{equation}
\label{eq:condition}
\begin{aligned}
    \hat{\epsilon} \leq \frac{R_{max}}{1-\gamma} - \kappa_{M,\gamma}&\bigg(\frac{\delta_{M,\gamma}/2}{1-\gamma_{Bw}(1-\delta_{M,\gamma}/2)} \\
    & \qquad + \frac{\hat{\delta}_{M,\gamma}/2}{1-\gamma_{Bw}(1-\hat{\delta}_{M,\gamma})} \bigg).
\end{aligned}
\end{equation}

\begin{figure}
\centering
\includegraphics[width = 0.47\textwidth]{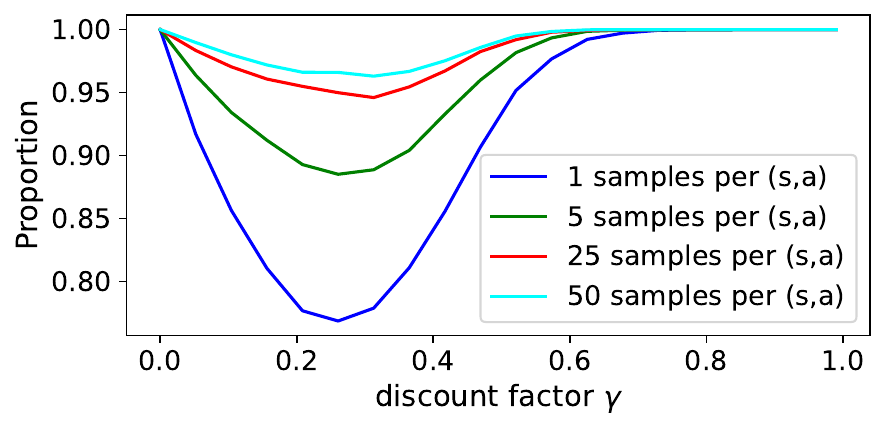}
\caption{Proportion of randomly sampled MDPs where Eq.~\ref{eq:condition} is true given a discount factor $\gamma$.}
\label{fig:gamma}
\end{figure}

Fig.~\ref{fig:gamma} supports the idea that Thm.~\ref{thm:planning_loss} becomes tighter than prior results when the variance due to model approximation (Def.~\ref{def:variancemodelApprox}) is low or when $\frac{R_{max}}{1-\gamma}$ is dominant ($\gamma$ close to $1$).

\section{Bias Under Partial Observability}
\label{sec:pomdp}


We now look at how partial observability impacts the structural parameters to better understand its impact on the bias. This is important since most practical problems suffer from a form of partial observability~\citep{dulac-arnold_challenges_2021}. 
%
%
We consider a discrete-time POMDP~\citep{sondik1978optimal}  described by the MDP tuple 
extended with two elements: a finite set of possible observations $\Omega$ and the probabilities of receiving each observation given a state, $\mathbb{O}$ : $\mathcal S \times \Omega \mapsto [0,1]$. On each time step $t\in\mathbb N_0$, the current state $S_t\in\mathcal S$ leads the agent to receive an observation $O_t \in \Omega$ (using $\mathbb{O})$, an action $A_t\in\mathcal A$ is played, the environment transitions into the next (unknown) state $S_{t+1}$ (using $P$) and generates an observed reward $R_{t+1}$ (using $R$). 

When facing a partially observable setting, an effective way of approximating a solution is to use a policy defined on compressed histories~\citep{francois-lavet_overfitting_2019}. Let $ \mathcal{H}_t = \Omega \times (\mathcal A \times R \times \Omega)^t$ denote the set of histories observed up to time $t$ and let $\mathcal{H} = \bigcup_{t=0}^\infty \mathcal{H}_t$ denote the space of all possible histories. The belief state $b(s|H)$ is a vector where the $i$-th component $(i \in \{1,\dots,|\mathcal S|\})$ corresponds to $\mathbb P(s = i | H)$, for history $H\in\mathcal H$. One can define a mapping $\phi : \mathcal{H} \mapsto \phi(\mathcal{H})$, where the set compressed history $\phi(\mathcal{H}) = \{ \phi(H) | H\in\mathcal H \}$ is finite, which can be used as input to a policy $\pi : \phi(\mathcal H) \mapsto \mathcal A$. We consider POMDPs with sufficient mappings such that $\mathbb P(s | H) = \mathbb P(s | \phi(H))$, which defines an MDP on state space $\phi(\mathcal{H})$.

Given a POMDP (extended tuple) $M$ and any given distribution $\mathcal{D}_H$ over histories, one can define the expected return obtained over an infinite time horizon from a given history $H$, with $A_t \sim \pi(\phi(H_t))$:\\
\begingroup
\fontsize{9}{8}
\begin{align*}
    V_{M,\gamma}^\pi(\phi(H)) &= \underset{\substack{H' \sim \mathcal{D}_H: \\ \phi(H') = \phi(H)}}{\mathbb{E} } \left[V_M^\pi(H'|\phi) \right]\\
\text{With} \ V_{M,\gamma}^\pi(H'|\phi) \ &\text{given by}\\
    V_{M,\gamma}^\pi(H'|\phi) &= \mathbb{E}_\pi \left[\sum_{k=0}^\infty \gamma^k R_{t+k+1} \Big| S_t \sim b(\cdot |H_t=H'), \right].
\end{align*}
\endgroup

For a given sufficient mapping $\phi$, the optimal policy $\pi_{M,\gamma}^*$ maximizes $V_{M,\gamma}^{\pi} (\phi(H))$ for all histories $H\in\mathcal H$.

\subsection{Extending Structural Parameters}

We can extend Definitions~\ref{def:value_function_variation} and~\ref{def:action_variation} to the POMDP setting by applying them to compressed histories rather than the actual states in the underlying MDP:
\begin{align}
\label{eq:value_function_variation_pomdp}
 \kappa_{M,\gamma}^\phi &= \max_{\sigma,\sigma' \in \phi(\mathcal{H})} \left| V^{\pi^*_{M,\gamma}}_{M,\gamma}(\sigma) - V^{\pi^*_{M,\gamma}}_{M,\gamma}(\sigma')  \right| \\
\label{eq:action_variation_pomdp}
       \delta_M^\phi &= \max_{\sigma \in \phi(\mathcal{H})} \max_{a,a' \in \mathcal{A}} \sum_{\sigma' \in \phi(\mathcal{H})} \left| \mathbb{P}(\sigma'|\sigma,a) - \mathbb{P}(\sigma'|\sigma,a') \right|.
\end{align}
%
We introduce the following result showing how the structural parameters in the POMDP relate to the structural parameters of the underlying MDP~(see Appendix F):
\begin{theorem}
\label{thm:struct_params_pomdp}
Given a POMDP $M$, let $ \kappa_{M,\gamma}^{\mathcal S} $ and $\delta_M^{\mathcal S}$ denote the structural parameters (Definitions~\ref{def:value_function_variation} and~\ref{def:action_variation}) evaluated on the underlying state space. Let $\mathcal{H}(s) = \bigcup_{t=0}^\infty \{H_t : b(s|H_t) > 0, H_t \in \mathcal{H}_t\}$ denote the set of all histories which can lead to being in state $s$ at any time $t$, and $\mathcal D_H(s)$ denote its probability distribution. Define $\Delta_{\epsilon}^{\phi} $ s.t. $\left| V^{\pi_{M,\gamma}^*}_{M,\gamma}(s) - V^{\pi_\phi}_{M,\gamma}(s) \right| \leq \Delta_{\epsilon}^{\phi}~ \forall s \in \mathcal S$, define \\ $\pi_{\phi}(s) = \underset{H \sim \mathcal{D}_{H}(s)}{\mathbb{E}} \pi^*_{M,\gamma}(\phi(H))$ as the optimal policy on compression of histories when executed over the underlying state space and define $b(s|\sigma) = \underset{H \sim \mathcal{D}_{H}(s), \sigma = \phi(H)}{\mathbb{E}} b(s|H)$ for $\sigma \in \phi(\mathcal{H})$. We have that:
    \begin{equation*}
    \begin{aligned}
            &\quad\quad\quad\quad\quad\quad \delta_M^\phi \leq \delta_M^{\mathcal S} \\[1em]
            \kappa_{M,\gamma}^\phi &\leq \max_{\sigma,\sigma' \in \phi(\mathcal{H})} \frac{\left\|  b( \cdot | \sigma) -  b( \cdot | \sigma') \right\|_1}{2} \left( \kappa_{M,\gamma}^{\mathcal{S}}  + \Delta_{\epsilon}^{\phi} \right).
    \end{aligned}
    \end{equation*}
\end{theorem}

The first inequality confirms that partial observability impacts negatively the ability for the agent to control state transitions. The second inequality implies that if the policy on the partially observable domain remains accurate on the underlying state space ($\Delta_{\epsilon}^{\phi} \approx 0$), then the state-value variation observed by the agent is lower (since the $L_1$ distance is bounded by $2$), which could make the learning task easier and as efficient with a low discount factor. This $L_1$ distance often has a value of $2$, which makes the bound quite loose, but they illustrate the idea that the values of structural parameters decrease when taking a convex combination over states. In fact the maximal value of the expectation of a random variable happens if all the mass is concentrated on the maximal value of the support. This is explained further in appendix B.
By considering that the horizon-sensitive action variation (Def.~\ref{def:horizonSensitiveActionVariation}) is upper-bounded by the action variation (Def.~\ref{def:action_variation}), we can observe from Eq.~\ref{eq:bias-extension} that the bias in the underlying MDP upper-bounds the bias of the POMDP when the optimal policy under partial observability is accurate on the true state space. This extends the ideas from~\citet{abel2016near} where abstractions can make a problem much easier to learn while retaining good performance and in our case, lower the bias.
We can get a better understanding of this trade-off in the state-abstraction setting~\citep{abel2016near}, as shown in our numerical experiments below.


\section{Numerical Experiments}
\label{sec:experiments}

We now conduct experiments\footnote{Experiments are conducted on a AMD Ryzen 5 3600 CPU and a GTX 1660 Ti GPU.
} to highlight the relationships between the planning horizon, the partial observability, and the structural parameters of the underlying MDP.

\paragraph{Random MDPs}
We consider the simulated environment of \citet{jiang2016structural}. We use 2-action MDPs, with $Fixed(|\mathcal{S}|,d)$ denoting a randomly generated MDP with $d\geq 1$ next states reachable from each state. MDPs are sampled using the following procedure: 1) each state-action pair is assigned $d$ possible next states; 2) transition probabilities to these states are sampled uniformly in $[0,1]$, then normalized; 3) rewards are assigned to state-action pairs by sampling uniformly in $[0,1]$.

\paragraph{Extension to Partial Observability} We consider the state-abstraction setting~\citep{abel2016near}, which corresponds to a specific case of partial observability where the history compressor $\phi(\mathcal{H})$ returns only the last observation and where $\mathbb{O}$ is a one-hot vector on an observation from $\Omega$. For simplicity, we make sure that each observation is connected to at least one state.
Using Bayes' theorem to recover the belief that the agent is in state $s$ given observation $\omega$, we get a constant uniform distribution on every state $s$ which maps onto this observation: 
\begin{align*}
    b(s|\omega) = \frac{1}{\lvert\{s \in \mathcal{S} : \mathbb{O}(o,s) > 0\} \rvert} \forall s \in \mathcal{S} : \mathbb{O}(\omega,s) > 0, \\ \forall \omega \in \Omega.
\end{align*}
and a belief of $0$ otherwise. From this special case of POMDP, we can extract an abstract MDP $M_A = \langle \mathcal{S}_A, \mathcal{A},P_A ,R_A, \gamma \rangle$ from the underlying MDP $M = \langle \mathcal{S}, \mathcal{A},P ,R, \gamma \rangle$~\citep{abel2016near}:
\begin{align}
        \label{eq:Abstractedparams1}
        R_A(\omega,a) &= \sum_{s \in  \mathcal{S}} b(s|\omega) R(s,a)
        \text{, and} \\
        \label{eq:Abstractedparams2}
        P_A(\omega,a,\omega') &= \sum_{s \in \mathcal{S}} \sum_{s' \in \mathcal{S}}P(s,a,s')b(s|\omega)\mathbb{O}(\omega',s').
\end{align}

\begin{figure*}[ht]
\centering
\includegraphics[width=0.45\textwidth]{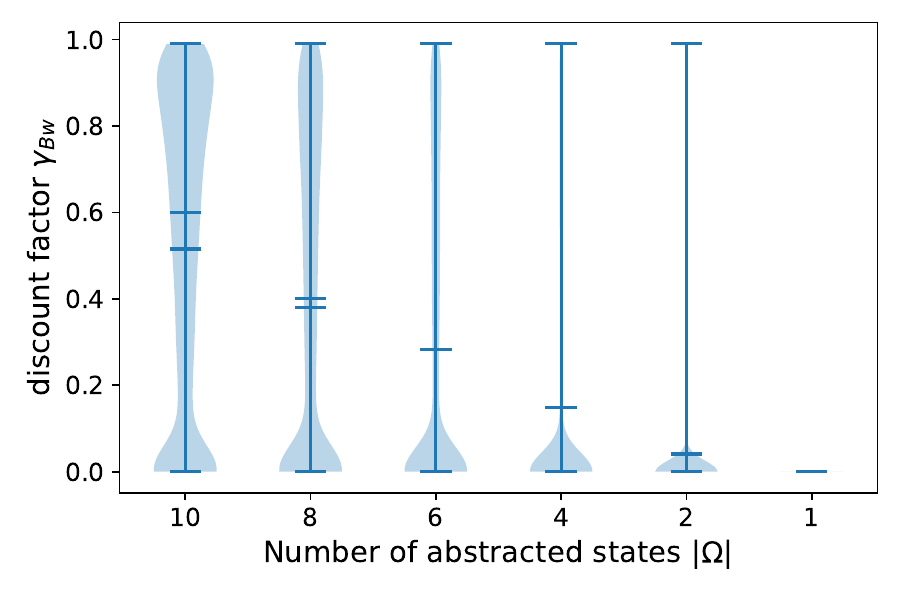}
\includegraphics[width=0.45\textwidth]{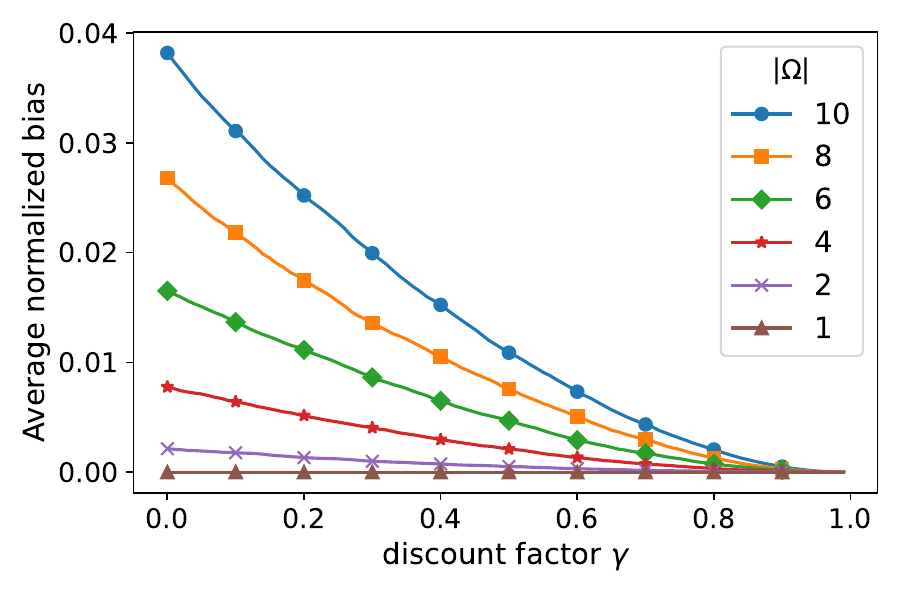}

\caption{Left: Distribution of Blackwell discount factors over $10^4$ POMDPs given the number of observations.  Right: Average normalized bias given the discount factor and number of observations.} 
\label{fig:results:bias}
\end{figure*}
For our experiments under partial observability, we start by sampling an MDP from $Fixed(|\mathcal{S}|,d)$. Then, we can map it onto abstracted MDPs (POMDPs) with different number of observations as $|\Omega|$. The number of obervations encodes the level of partial observability akin to what happens in discretization. For $|\Omega| = |\mathcal S|$, the problem is fully observable. For $|\Omega| = 1$ (and $|\mathcal S| > 1$), the agent is completely blind to the state. We sample $10^4$ MDPs with $Fixed(10,3)$ and abstract each MDP into 6 configurations: $|\Omega| \in \{ 10, 8, 6, 4, 2, 1 \}$. The Blackwell discount factors are computed by iterating from $\gamma = 1$ to $\gamma = 0$ with step size of $0.01$ until the optimal policy changes.

Fig.~\ref{fig:results:bias} (left) shows that the mass of the Blackwell planning horizon tends to decrease as the observability decreases.
Since the bias is null when planning with a discount factor larger than $\gamma_{Bw}$, we only cumulate variance above that point. 
When $|\Omega|=1$, the myopic agent ($\gamma=0$) enjoys the optimal planning horizon, which corresponds to the bandit setting. We also evaluate the normalized bias\\
$$
    \max_{s \in \mathcal{S}} \left(V_{M,\gamma_{Bw}}^{\pi_{M,\gamma_{Bw}}^{*}}(s) - V_{M,\gamma_{Bw}}^{\pi_{M, \gamma}^{*}} (s) \right)/  V_{M,\gamma_{Bw}}^{\pi_{M,\gamma_{Bw}}^{*}}(s)
$$\\
for different planning horizons, averaged over different levels of partial observability. Fig.~\ref{fig:results:bias} (right) shows that, although the bias decreases when the planning horizon ($1/(1-\gamma)$) increases, this effect attenuates as the observability decreases.
Since many real-world problems are partially observable, this supports the need to consider shallow planning more seriously.


\begin{figure*}[ht]
\centering
\includegraphics[width=0.45\textwidth]{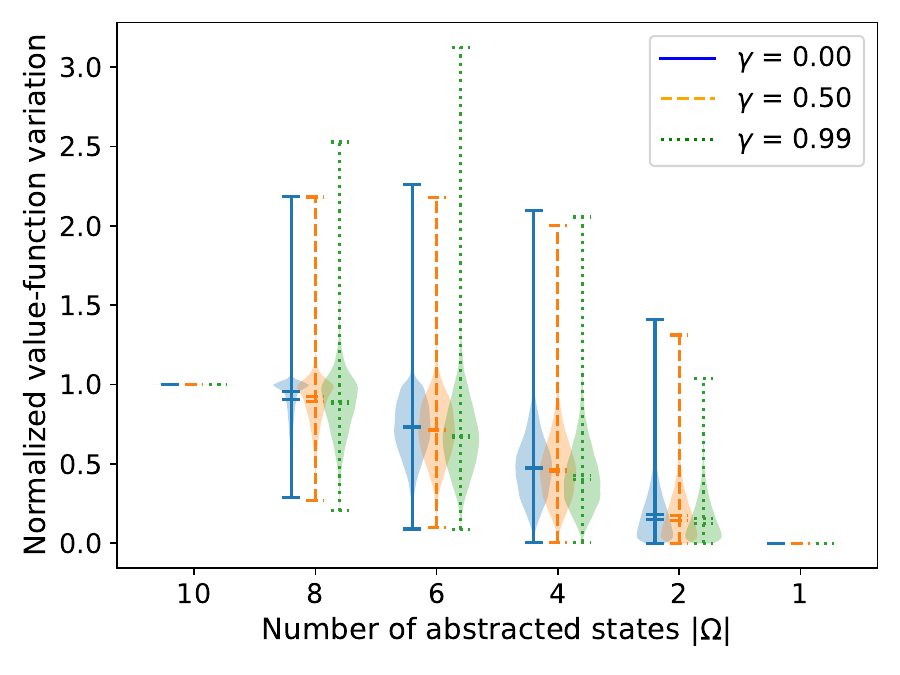}
\includegraphics[width=0.45\textwidth]{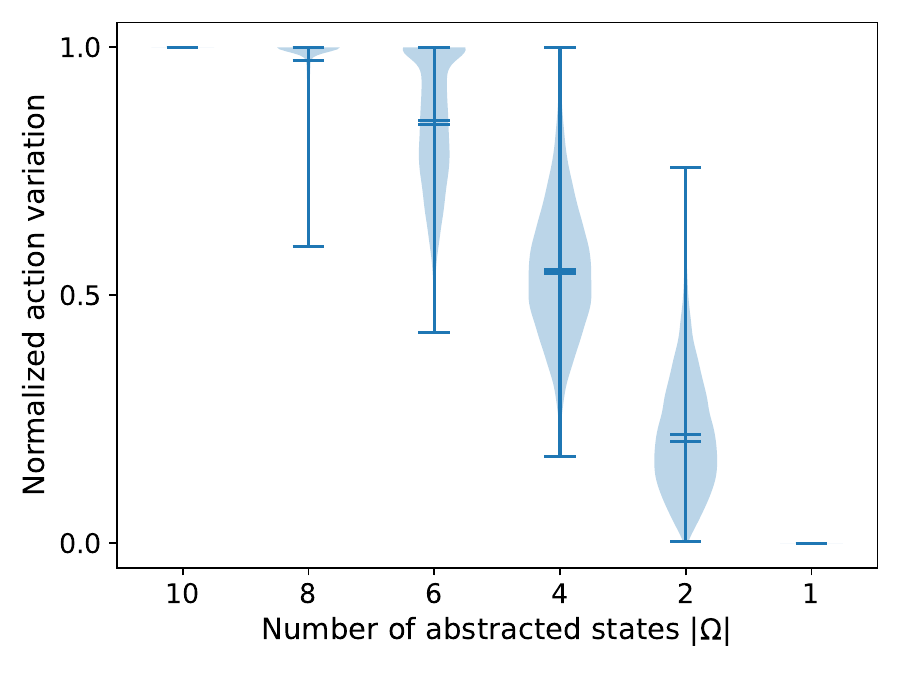}
\caption{Left: Distribution of normalized $\kappa_{M,\gamma}^\phi$ over $10^4$ POMDPs given the number of observations and the discount factor used. Right: Distribution of normalized $\delta_M^\phi$ given the number of observations.} 
\label{fig:results:params}
\end{figure*}

We normalize parameters for each abstract MDP by dividing structural parameters by those of the true $Fixed(\cdot)$ MDP, i.e., $\kappa_{M,\gamma}^\phi/\kappa_{M,\gamma}^S$ for the normalized value function variation and $\delta_M^\phi/\delta_M^S$ for the normalized action variation.\\

Fig.~\ref{fig:results:params} (left) highlights the tradeoff given by Thm.~\ref{thm:struct_params_pomdp}. We see that there is a downards trend, but for high values of gamma, the error $\Delta_{\epsilon}^{\phi}$ in some cases is so high that the value-function variation can be higher under partial observability.  Fig.~\ref{fig:results:params} (right) highlights the strict inequality offered by Thm.~\ref{thm:struct_params_pomdp} on the action variation of the POMDP vs the underlying MDP. By observing Thm.~\ref{thm:planning_loss} and Fig.~\ref{fig:results:params}, the reduction in structural parameters offers insight into why the bias decreases under partial observability. It also points to the fact that there might exist a bound on the Blackwell discount factor using structural parameters much tighter than \citet{gr-Clément_Petrik_2023} or even improve upon the results of \citet{laidlaw2023bridging} using partial observability.  


\section{Impact of Partial Observability on Deep RL}


We explore the interaction of shallow planning and partial observability in the Cartpole-v1 environment~\citep{towers_2024_13902792}. In Cartpole, an agent uses two discrete actions (left or right) to balance a pole on a moving cart. The agent's state includes the cart's position and velocity, along with the pole's angle and angular velocity. A reward of $+1$ is given for each time step before episode termination or reaching 500 time steps.

We consider the widely used PPO~\cite{schulman2017proximal} agent policy with the recommended hyperparameters for this task~\cite{stable-baselines3}. We then consider different discount factors $\gamma \in \{0,0.24,0.49,0.73,0.98\}$, with the largest $\gamma = 0.98$ from the baseline~\cite{stable-baselines3}. For the partially observable component, we simulate noisy sensors, which are common in real life. These are simulated by injecting noise into the state. The noise is sampled from a multivariate normal distribution $\mathcal{N}(\mathbf{0},I \sigma^2)$ parameterized by a diagonal covariance matrix with value $\sigma^2$ on the diagonal. We consider $\sigma \in \{ 0, 0.1, 1 \}$. We train 10 agents for each combination of $(\gamma, \sigma)$, resulting in 150 models, and evaluate each of these models on 100 unseen seeds.\\


\begin{figure}
    \centering
    \includegraphics[width=0.48\textwidth]{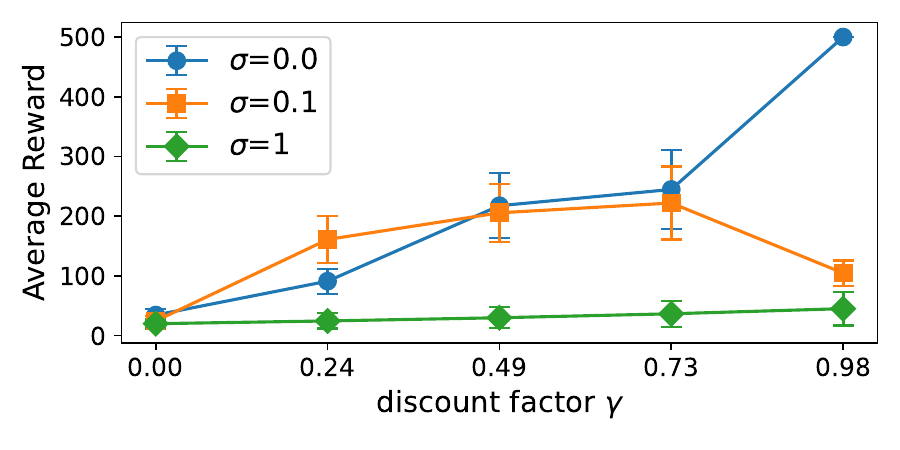}
    \caption{Average reward and standard deviation obtained by running 10 models on 100 environment seeds given the noise level and discount factor.}
    \label{fig:result:cartpole}
\end{figure}

Fig.~\ref{fig:result:cartpole} displays the average reward obtained by running the 10 models associated with each $(\gamma, \sigma)$ configuration on the 100 environment seeds. Without surprise, planning with the longest horizon (largest discount factor) maximizes rewards (as expected on this task). However, we also observe that, as the environment becomes partially observable (e.g. $\sigma=0.1$), shallow planning can be better than long planning since the bias is lowered from the partial observability. The Blackwell discount factor could be lowered, supporting our previous results and thesis. As expected, when observability becomes too low (e.g. $\sigma=1$), the agent cannot understand the environment anymore and is unable to learn the task no matter the discount factor considered.


\section{Related Work}
\label{sec:related_work}

\paragraph{Effective Planning Horizon} When it comes to using RL in practical applications, there is a trend towards a better understanding of the planning horizon on the optimality of the policy learned. \citet{gr-Clément_Petrik_2023} explore the Blackwell optimality criterion and proves the existence of a planning horizon above which the policy learned cannot be improved. Similarly, \citet{laidlaw2023bridging} explore the minimal horizon (in number of time steps forward) for which the optimal policy can be retrieved with high probability. Our work builds on top of these frameworks to better understand the bias-variance trade-off that arises from using a shallow planning horizon.

\paragraph{Planning Loss} The proposed decomposition of the planning loss (Eq.~\ref{eq:planning_loss}) enables its interpretation in terms of bias and variance, unlike the previous decomposition of \citet{jiang2015dependence}. The bias term had been studied previously~\cite{jiang2016structural}, borrowing inspiration on the approximation term from the PAC bound literature in machine learning~\citep{shalev-shwartz_understanding_2014}. However, the variance term remained unbounded. By connecting ideas from previous works \citet{jiang2015dependence,jiang2016structural}, we are now able to understand not only the bias as a function of the structural parameters, but also part of the variance.

\paragraph{Partial Observability}

Most practical problem are under some form of partial observability~\citep{dulac-arnold_challenges_2021}. Different approximation schemes are explored through the literature like memoryless policies~\citep{muller2021geometry} or compressions of histories~\citep{francois-lavet_overfitting_2019}. There is also plenty of work on properties of POMDPs such that approximations are more likely to yield good results \citep{Liu_Chung_Szepesvári_Jin_2022}. Our work extends these results to a better understanding of how these approximation schemes can impact not only the sample complexity or the performance guarantees, but the planning horizon that should be considered.

In this direction, the literature on state abstraction aims to tackle some of the challenges of practical RL through a simpler state-space~\citep{abel2016near,Allen_Parikh_Gottesman_Konidaris_2021,Abel_Arumugam_Lehnert_Littman_2018}. This approximation scheme is well understood, and lots of useful properties on policy performance or sample complexity have been found. These results have not yet been connected to the planning horizon that should be used under these regimes. We attempt to bridge this gap by providing insight into the behaviour of the bias under partial observability.

\section{Conclusion}


We extend existing structural parameters to consider the planning horizon~(Def.~\ref{def:horizonSensitiveActionVariation}) and the model approximation (Def.~\ref{def:empirical_action_variation}). This allow us to extend an existing bound on the bias (Eq.~\ref{eq:bias-extension}) and propose a new bound on the variance (Lem.~\ref{lemma:variance}), which result in a new bound on the planning loss (Thm.~\ref{thm:planning_loss}). We finally extend the structural parameters to POMDPs (Eq.~\ref{eq:value_function_variation_pomdp} and~\ref{eq:action_variation_pomdp}) and show that these are controlled by their fully observable counterparts (Thm.~\ref{thm:struct_params_pomdp}). This complements previous results~\citep{abel2016near,francois-lavet_overfitting_2019} by considering the impact of the planning horizon on the bias when shallow planning under partial observability.

\paragraph{Limitations and Future Work} Our work motivates further research on understanding when shallow planning might be beneficial by providing new theory. 
However, our results are limited to a finite state space. Also, the reduction in $\kappa_{M,\gamma}^\mathcal{S}$~(Thm.~\ref{thm:struct_params_pomdp}) that comes from partial observability can be much larger than what is captured by the $L_1$ distance; the bound could be improved by using better mathematical tools. Even with these limitations, practitioners can use our analysis and our experiments to better understand how the nature of their problem might impact the bias-variance trade-off, and select a better planning horizon. Future work could improve upon our results by using deep-learning theory. It could also be possible to develop automatic discount factor selection algorithms, as previous research papers have indicated that even naive form of this strategy can be very impactful~\citep{franccois2015discount}. State-specific discount factors are also interesting. It is easy to construct an MDP such that the Blackwell discount factor is very high because of only a handful of states in the state space. A better understanding of the local structure around a state means we could create heuristics to adapt the discount factor automatically~\citep{pitis2019rethinking,yoshida2013reinforcement}. Finally, a better understanding of the process which could lead to the downward trend in $\gamma_{Bw}$ observed in Fig.~\ref{fig:results:bias} through our analysis could provide better insight into how to prevent over-fitting.



\section*{Acknowledgements}

We acknowledge financial support from Intact and CIFAR.

\bibliography{aaai25}

\clearpage
\onecolumn

\appendix

\section{Appendix A: Notation}
\label{app:notation}
\begin{table}[h]
    \centering
    \small
    \begin{tabular}{|c|l|}
        \hline
        \textbf{Symbol} & \textbf{Description} \\
        \hline
        \( \mathcal{S} \) & Finite state space \\
        \( \mathcal{A} \) & Finite action space \\
        \( P \) & Transition function \( P: \mathcal{S} \times \mathcal{A} \times \mathcal{S} \mapsto [0,1] \) \\
        \( R \) & Reward function \( R: \mathcal{S} \times \mathcal{A} \mapsto [0, R_{\text{max}}] \) \\
        \( \gamma_{Bw} \) & Blackwell discount factor \\
        \( V_{M,\gamma}^\pi \) & Vector of state values when following policy $\pi$ on MDP $M$ with discount factor $\gamma$ \\

        \( \pi^*_{M,\gamma} \) & Optimal policy in MDP \( M \) with discount factor \( \gamma \) \\

        \( \mathcal{Z}_{M}(\pi \neq \pi') \) & Set of state-action pairs on MDP $M$ where two policies \( \pi \) and \( \pi' \) differ \\
        \( \hat{\epsilon} \) & Variance due to model approximation \\
        \( \delta_{M} \) &  Action variation for MDP $M$ \\
        \( \kappa_{M,\gamma} \) & Value-function variation in MDP $M$ with discount factor $\gamma$ \\
        \( \delta_{M,\gamma} \) & Horizon-sensitive action variation in MDP $M$ with discount factor $\gamma$ \\
        \( \hat{\delta}_{M,\gamma} \) & Empirical action variation in MDP $M$ with discount factor $\gamma$ \\

        \( P^{\pi}_{s,k} \) &  $|\mathcal{S}| \times |\mathcal{S}|$ matrix of transition probabilities from state \( s \) when following policy \( \pi \) for \( k \) steps \\
        $[P^\pi_M]$ & $|\mathcal{S}| \times |\mathcal{S}|$ matrix of transition probabilities when following policy $\pi$ in MDP $M$ \\
        $e_{s}^{\top}$ & One-hot vector of length $|\mathcal S|$ with a $1$ at the index of $s \in \mathcal{S}$\\
        \hline
        \( \Omega \) & Finite set of observations \\
        \( \mathbb{O} \) & Observation probabilities \( \mathbb{O}: \mathcal{S} \times \Omega \mapsto [0,1] \) \\
        \( \mathcal{H}_t \) & Set of histories observed up to time \( t \) \\
        \( \mathcal{H} \) & Space of all possible histories \\
        \( \mathcal{D}_H \) & Distribution over histories \\
        \( b(s|H) \) & Belief state, \( \mathbb{P}(s = i | H) \) for history \( H \in \mathcal{H}\) \\
        \( \phi \) & Mapping from histories to a finite set, \( \phi: \mathcal{H} \mapsto \phi(\mathcal{H}) \) \\
        $\phi(\mathcal{H})$ & Space of all possible compression of histories \\
        \( \pi(\phi(H)) \) & Policy defined on compressed histories \\
        $\pi_{\phi}(s)$ & Optimal policy defined over compressed histories executed on the true state space \\
        \( V_{M,\gamma}^\pi(\phi(H)) \) & Expected return from history \( H \) with policy \( \pi \) \\
        \( \pi_{M,\gamma}^* \) & Optimal policy for POMDP \( M \) and discount factor \( \gamma \) \\
        $\Delta_{\epsilon}^{\phi}$ & Loss of performance on the true state space from executing $\pi_{\phi}(s)$ instead of the optimal policy \\
        \( \kappa_{M,\gamma}^\phi \) & Value-function variation in POMDP $M$ with discount factor $\gamma$ and mapping $\phi$ \\
        $\delta_M^\phi$ & Action variation in POMDP $M$ and mapping $\phi$ \\
        \hline
    \end{tabular}
    \vspace{0.1in}
    \caption{List of notations}
    \label{tab:notations}
\end{table}

\clearpage

\section{Appendix B: Proof of Eq. \ref{eq:bias-extension}}
\label{app:proof_bias}

Let $[P^\pi_M]$ denote a $|\mathcal{S}| \times |\mathcal{S}|$ matrix with the element indexed by $(s,s')$ being $P(s'|s,a)$ under policy $\pi$ and $e_{s}^{\top}$ be a one-hot vector on an arbitrary state $s \in \mathcal{S}$, that is a vector of $0s$ and a value of $1$ for the index $s$. We use the following result: 

\begin{fact}[State value decomposition from \citet{jiang2016structural}]
\label{Fact:DecompositionGamma}
Let $M$ be and MDP with Blackwell discount factor $\gamma_{Bw} > \gamma$:
\begin{equation*}
    \begin{aligned}
    V_{M,\gamma_{Bw}}^{\pi}(s)
    &=e_{s}^{\top}V_{M,\gamma}^{\pi}+\sum_{k=1}^{\infty}(\gamma_{Bw}-\gamma)\gamma_{Bw}^{k-1}e_{s}^{\top}[P_M^{\pi}]^{k}V_{M,\gamma}^{\pi}.
    \end{aligned}
\end{equation*}
\end{fact}

We then have for any arbitrary state $s \in \mathcal{S}$ by using fact \ref{Fact:DecompositionGamma} on both state values of the bias:
\begin{align*}
    V_{M,\gamma_{Bw}}^{\pi_{M,\gamma_{Bw}}^{*}} (s) - V_{M,\gamma_{Bw}}^{\pi_{M, \gamma}^{*}} (s) &= e_{s}^{\top}\valueState{M,\gamma}{\pi_{M,\gamma_{Bw}}^{*}} - e_{s}^{\top}\valueState{M,\gamma}{\pi_{M,\gamma}^{*}} \\
 & \qquad + (\gamma_{Bw}-\gamma)\sum_{k=1}^{\infty}\gamma_{Bw}^{k-1} \left( e_{s}^{\top}[P_M^{\pi_{M,\gamma_{Bw}}}]^{k}V_{M,\gamma}^{\pi^*_{M,\gamma_{Bw}}} - \ets\transitionMatrix{M}{\pi_{M,\gamma}}^k\valueState{M,\gamma}{\pi^*_{M,\gamma}} \right).\\
 & \leq 0 + (\gamma_{Bw}-\gamma)\sum_{k=1}^{\infty}\gamma_{Bw}^{k-1} \left( e_{s}^{\top}[P_M^{\pi_{M,\gamma_{Bw}}}]^{k}V_{M,\gamma}^{\pi^*_{M,\gamma}} - \ets\transitionMatrix{M}{\pi_{M,\gamma}}^k\valueState{M,\gamma}{\pi^*_{M,\gamma}} \right).
\end{align*}

By realizing that $\valueState{M,\gamma}{\pi_{M,\gamma}^{*}} (s) \geq \valueState{M,\gamma}{\pi_{M,\gamma_{Bw}}^{*}} (s) \ \forall s \in \mathcal{S}$, the first term is upper bounded by $0$. We can also start by bounding the inner term of the sum. To achieve this, we will use the following result:
 \begin{lemma}[\citet{jiang2016structural}]
\label{lemma2}
    Given stochastic vectors $p,q \in \mathbb{R}^{|S|}$, and a real vector $V$ with the same dimension:
    $$|p^\top V-q^\top V|\leq\|p-q\|_1\max_{s,s^\prime}|V(s)-V(s^\prime)|/2.$$
\end{lemma}

In a way, this lemma gives a bound on the difference between two expectations defined over the same support. The result is quite loose since it doesn't take well into consideration the form of the distribution in $p$ and $q$. We will still use it since its the only result we know, and will still be useful to illustrate our ideas.

 We can bound the inner sum using Lemma~\ref{lemma2}:
 \begin{align*}
     e_{s}^{\top}[P_M^{\pi_{M,\gamma_{Bw}}}]^{k}V_{M,\gamma}^{\pi^*_{M,\gamma}} - \ets\transitionMatrix{M}{\pi_{M,\gamma}}^k\valueState{M,\gamma}{\pi^*_{M,\gamma}} \leq \left\| \ets \transitionMatrix{M}{\pi_{M,\gamma_{Bw}}}^{k} - \ets \transitionMatrix{M}{\pi_{M,\gamma}}^{k} \right\|_{1}  \frac{\kappa_{M,\gamma}}{2}.
 \end{align*}

In their proof, \citet{jiang2016structural} use the action variation as an upper bound to the row-wise $L_1$-norm distance between the transition matrices $\transitionMatrix{M}{\pi_1}$ and $\transitionMatrix{M}{\pi_2}$ of two policies $\pi_1$ and $\pi_2$. By realizing that this distance is $0$ on states where $\pi_1(s) = \pi_2(s)$, we can tighten up the bounds using Def.~\ref{def:discordantSApair}, changing a single step in the proof of Lemma 1 from \citet{jiang2016structural}, and using their Corollary 4 with our new maximal $l_1$-norm distance in definition~\ref{def:horizonSensitiveActionVariation} and definition~\ref{def:empirical_action_variation} to get the following inequalities, which are equivalent to Propositions~\ref{prop:l1dist:horizon_sensitive} and~\ref{prop:l1dist:empirical} respectively: 
\begin{equation*}
    \begin{aligned}
        \left\| \ets \transitionMatrix{M}{\pi^*_{M,\gamma}}^{k} - \ets \transitionMatrix{M}{\pi^*_{M,\gamma_{Bw}}}^{k} \right\|_{1} &\leq 2 - 2(1- \delta_{M,\gamma}/2)^k\\
        \left\| \ets \transitionMatrix{M}{\pi^*_{M,\gamma}}^{k} - \ets \transitionMatrix{M}{\pi^*_{\wh{M},\gamma}}^{k} \right\|_{1} &\leq 2 - 2(1-\hat \delta_{M,\gamma}/2)^k.
    \end{aligned}
\end{equation*}

By plugging the second inequality in our inner term, we obtain the following bound:
\begin{align*}
    V_{M,\gamma_{Bw}}^{\pi_{M,\gamma_{Bw}}^{*}} (s) - V_{M,\gamma_{Bw}}^{\pi_{M, \gamma}^{*}} (s) &\leq  \sum_{k=1}^\infty(\gamma_{Bw}-\gamma)\gamma_{Bw}{}^{k-1}\big(1-(1-\delta_{M,\gamma}/2)^k\big)\kappa_{M,\gamma} \\
    &= \frac{\delta_{M,\gamma}/2\cdot\kappa_{M,\gamma}(\gamma_{Bw}-\gamma)}{(1-\gamma_{Bw})(1-\gamma_{Bw}(1-\delta_{M,\gamma}/2))}.
\end{align*}
\hfill\ensuremath{\square}

\section{Appendix C: Proof of Lemma~\ref{lemma:variance}}
\label{app:proof_variance}

For this proof, we will use the same setting as section \ref{app:proof_bias}. Using Fact~\ref{Fact:DecompositionGamma} on both state values in the variance, we have the following:
\begin{equation}
\begin{aligned}
\label{eq:varianceDecomp}
\underbrace{V_{M,\gamma_{Bw}}^{\pi_{M,\gamma}^{*}}(s) - V_{M,\gamma_{Bw}}^{\pi_{\wh{M}, \gamma}^{*}}(s)}_{\text{variance}}
&=e_{s}^{\top}\valueState{M,\gamma}{\pi_{M,\gamma}^{*}} - e_{s}^{\top}\valueState{M,\gamma}{\pi_{\wh{M},\gamma}^{*}}\\[-15pt]
&\qquad + (\gamma_{Bw}-\gamma)\sum_{k=1}^{\infty}\gamma_{Bw}^{k-1} \left( e_{s}^{\top}[P_M^{\pi_{M,\gamma}}]^{k}V_{M,\gamma}^{\pi^*_{M,\gamma}} - \ets\transitionMatrix{M}{\pi_{\wh{M},\gamma}}^k\valueState{M,\gamma}{\pi^*_{\wh{M},\gamma}} \right).
\end{aligned}
\end{equation}

We will start by bounding the sum first. The inner term can be rewritten as:
\begin{equation}
    \begin{aligned}
    \label{eq:decompInnerSumEarly}
    \ets \transitionMatrix{M}{\pi_{M,\gamma}}^{k} \valueState{M,\gamma}{\pi^*_{M,\gamma}} 
    - \ets \transitionMatrix{M}{\pi_{\wh{M},\gamma}}^{k} \valueState{M,\gamma}{\pi^*_{\wh{M},\gamma}} &=
    \ets \transitionMatrix{M}{\pi_{M,\gamma}}^{k} \valueState{M,\gamma}{\pi^*_{M,\gamma}} 
    -\ets \transitionMatrix{M}{\pi_{\wh{M},\gamma}}^{k} \valueState{M,\gamma}{\pi^*_{M,\gamma}}\\
    &+   \ets \transitionMatrix{M}{\pi_{\wh{M},\gamma}}^{k} \valueState{M,\gamma}{\pi^*_{M,\gamma}} - \ets \transitionMatrix{M}{\pi_{\wh{M},\gamma}}^{k} \valueState{M,\gamma}{\pi^*_{\wh{M},\gamma}} .
    \end{aligned}
\end{equation}

Note that there is a loss in tightness that comes from the fact that to isolate $\hat{\epsilon}$ (to use the existing litterature on the variance term), we need to use this decomposition, on which we will then use Holder's inequality. For the next results, we use Lemma~\ref{lemma2} on the first difference, and using Holder's inequality on the second, we obtain:
\begin{equation}
\label{eq:decompositionInnerTerm}
    \begin{aligned}
        \ets \transitionMatrix{M}{\pi_{M,\gamma}}^{k} \valueState{M,\gamma}{\pi^*_{M,\gamma}} 
    - \ets \transitionMatrix{M}{\pi_{\wh{M},\gamma}}^{k} \valueState{M,\gamma}{\pi^*_{M,\gamma}} &\leq \left\| \ets \transitionMatrix{M}{\pi_{M,\gamma}}^{k} - \ets \transitionMatrix{M}{\pi_{\wh{M},\gamma}}^{k} \right\|_{1}  \frac{\kappa_{M,\gamma}}{2} \\
    &\quad + \left\|\valueState{M,\gamma}{\pi^*_{M,\gamma}} - \valueState{M,\gamma}{\pi^*_{\wh{M},\gamma}} \right\|_{\infty}\\
    &= \left\| \ets \transitionMatrix{M}{\pi_{M,\gamma}}^{k} - \ets \transitionMatrix{M}{\pi_{\wh{M},\gamma}}^{k} \right\|_{1}  \frac{\kappa_{M,\gamma}}{2} + \hat{\epsilon}.
    \end{aligned}
\end{equation}


For bounding the first term, we use the inequality presented in section \ref{app:proof_bias}.Then, we have that the term inside the sum from Eq~\ref{eq:varianceDecomp} is bounded by: 
\begin{equation*}
    \begin{aligned}
        \ets \transitionMatrix{M}{\pi_{M,\gamma}}^{k} \valueState{M,\gamma}{\pi^*_{M,\gamma}} 
    - \ets \transitionMatrix{M}{\pi_{\wh{M},\gamma}}^{k} \valueState{M,\gamma}{\pi^*_{\wh{M},\gamma}} \leq \frac{(2 - 2(1-\hat \delta_{M,\gamma}/2)^k)\kappa_{M,\gamma}}{2} + \hat \epsilon.
    \end{aligned}
\end{equation*}

This allows us to obtain the following bound on the variance:
\begin{equation*}
    \begin{aligned}
         \underbrace{V_{M,\gamma_{Bw}}^{\pi_{M,\gamma}^{*}}(s) - V_{M,\gamma_{Bw}}^{\pi_{\wh{M}, \gamma}^{*}}(s)}_{\text{variance}} &\leq \hat \epsilon + \sum_{k=1}^\infty(\gamma_{Bw}-\gamma)\gamma_{Bw}{}^{k-1}\big(1-(1-\hat \delta_{M,\gamma}/2)^k\big)\kappa_{M,\gamma} + \frac{(\gamma_{Bw} -\gamma) \hat \epsilon}{1-\gamma_{Bw}} \\
         &= \hat \epsilon\left(\frac{1-\gamma}{1-\gamma_{Bw}}\right) + \frac{\hat \delta_{M,\gamma}/2\cdot\kappa_{M,\gamma}(\gamma_{Bw}-\gamma)}{(1-\gamma_{Bw})(1-\gamma_{Bw}(1-\hat \delta_{M,\gamma}/2))}.
    \end{aligned}
\end{equation*}

\hfill\ensuremath{\square}

\section{Appendix D: Proof of Theorem 1}
\label{app:proof_planning_loss}



Recall the following decomposition of the planning loss (Eq.~\ref{eq:planning_loss}):
\begin{align*}
\lVert V_{M,\gamma_{Bw}}^{\pi_{M,\gamma_{Bw}}^{*}} -V_{M,\gamma_{Bw}}^{\pi_{\wh{M},\gamma}^{*}} \rVert_{\infty}
&\leq  \underbrace{ \lvert\lvert V_{M,\gamma_{Bw}}^{\pi_{M,\gamma_{Bw}}^{*}} - V_{M,\gamma_{Bw}}^{\pi_{M, \gamma}^{*}} \rvert\rvert_{\infty} }_{\text{bias}}
+ \underbrace{ \lvert\lvert V_{M,\gamma_{\text{Bw}}}^{\pi_{M,\gamma}^{*}} - V_{M,\gamma_{Bw}}^{\pi_{\wh{M}, \gamma}^{*}} \rvert\rvert_{\infty}}_{\text{variance}}.
\end{align*}

We can upper-bound the bias using a result from \citet{jiang2016structural}, which we extended (see Appendix~\ref{app:BiasExtension}) to explicit the dependence on the planning horizon using Def.~\ref{def:horizonSensitiveActionVariation}:
\begin{align*}
    \lvert \lvert \underbrace{V_{M,\gamma_{Bw}}^{\pi_{M,\gamma_{Bw}}^{*}} - V_{M,\gamma_{Bw}}^{\pi_{M, \gamma}^{*}}}_{\text{bias}} \rvert \rvert_{\infty} \leq \frac{\delta_{M,\gamma}/2\cdot\kappa_{M,\gamma}(\gamma_{Bw}-\gamma)}{(1-\gamma_{Bw})(1-\gamma_{Bw}(1-\delta_{M,\gamma}/2))}.
\end{align*}



We can combine the bias upper bound with Lemma~\ref{lemma:variance} and obtain Thm.~\ref{thm:planning_loss}:
\begin{align*}
    \lVert V_{M,\gamma_{Bw}}^{\pi_{M,\gamma_{Bw}}^{*}} -V_{M,\gamma_{Bw}}^{\pi_{\wh{M},\gamma}^{*}} \rVert_{\infty}
&\leq  \underbrace{ \lvert\lvert V_{M,\gamma_{Bw}}^{\pi_{M,\gamma_{Bw}}^{*}} - V_{M,\gamma_{Bw}}^{\pi_{M, \gamma}^{*}} \rvert\rvert_{\infty} }_{\text{bias}}
+ \underbrace{ \lvert\lvert V_{M,\gamma_{\text{Bw}}}^{\pi_{M,\gamma}^{*}} - V_{M,\gamma_{Bw}}^{\pi_{\wh{M}, \gamma}^{*}} \rvert\rvert_{\infty}}_{\text{variance}}. \\
&\leq \frac{\delta_{M,\gamma}/2\cdot\kappa_{M,\gamma}(\gamma_{Bw}-\gamma)}{(1-\gamma_{Bw})(1-\gamma_{Bw}(1-\delta_{M,\gamma}/2))}
+ \hat \epsilon\left(\frac{1-\gamma}{1-\gamma_{Bw}}\right) + \frac{\hat \delta_{M,\gamma}/2\cdot\kappa_{M,\gamma}(\gamma_{Bw}-\gamma)}{(1-\gamma_{Bw})(1-\gamma_{Bw}(1-\hat \delta_{M,\gamma}/2))} \\
&= \kappa_{M,\gamma}\left(\frac{\gamma_{Bw} - \gamma}{1-\gamma_{Bw}}\right)\left( \frac{\delta_{M,\gamma}/2}{1-\gamma_{Bw}(1-\delta_{M,\gamma}/2)} +  \frac{\hat{\delta}_{M,\gamma}/2}{1-\gamma_{Bw}(1-\hat{\delta}_{M,\gamma}/2)} \right) + \hat \epsilon\left(\frac{1-\gamma}{1-\gamma_{Bw}}\right),
\end{align*}
where the last equality is obtained by rearranging the terms.

\section{Appendix E: Tightness of Theorem~\ref{thm:planning_loss}}
\label{app:tightness}
 
%
%
The bound offered by Thm.~\ref{thm:planning_loss} is tighter than prior results~\citep{jiang2015dependence} when the gain in tightness from using structural parameters is higher than the loss incurred from having a looser variance term in Eq.~\ref{eq:planning_loss}. Formally, we are looking for the condition such that Thm.~\ref{thm:planning_loss} is tighter than prior results~\citep{jiang2015dependence}:
\begin{align*}
\label{eq:comparison}
        \kappa_{M,\gamma}\left(\frac{\gamma_{Bw} - \gamma}{1-\gamma_{Bw}}\right)\left( \frac{\delta_{M,\gamma}/2}{1-\gamma_{Bw}(1-\delta_{M,\gamma}/2)} +  \frac{\hat{\delta}_{M,\gamma}/2}{1-\gamma_{Bw}(1-\hat{\delta}_{M,\gamma}/2)} \right) + \hat \epsilon\left(\frac{1-\gamma}{1-\gamma_{Bw}}\right) \leq& \\
        \frac{\gamma_{Bw}-\gamma}{(1-\gamma_{Bw})(1-\gamma)}R_{\max} + \hat{\epsilon}&,
\end{align*}
which we can re-arrange until we obtain Eq.~\ref{eq:condition}:
\begin{align*}
           &\kappa_{M,\gamma}\left( \frac{\delta_{M,\gamma}/2}{1-\gamma_{Bw}(1-\delta_{M,\gamma}/2)} +  \frac{\hat{\delta}_{M,\gamma}/2}{1-\gamma_{Bw}(1-\hat{\delta}_{M,\gamma}/2)} \right) + \hat \epsilon\left(\frac{1-\gamma}{\gamma_{Bw} - \gamma}\right) \leq\frac{R_{\max}}{(1-\gamma)}+ \hat{\epsilon}\left(\frac{1-\gamma_{Bw}}{\gamma_{Bw}-\gamma} \right)\\
          &\kappa_{M,\gamma}\left( \frac{\delta_{M,\gamma}/2}{1-\gamma_{Bw}(1-\delta_{M,\gamma}/2)} +  \frac{\hat{\delta}_{M,\gamma}/2}{1-\gamma_{Bw}(1-\hat{\delta}_{M,\gamma}/2)} \right) - \frac{R_{\max}}{(1-\gamma)}  \leq  \hat{\epsilon}\left(\frac{1-\gamma_{Bw}}{\gamma_{Bw}-\gamma} \right) - \hat \epsilon\left(\frac{1-\gamma}{\gamma_{Bw} - \gamma}\right) \\
          &\hat{\epsilon} \leq \frac{R_{max}}{1-\gamma}- \kappa_{M,\gamma}\left(\frac{\delta_{M,\gamma}/2}{1-\gamma_{Bw}(1-\delta_{M,\gamma}/2)} + \frac{\hat{\delta}_{M,\gamma}/2}{1-\gamma_{Bw}(1-\hat{\delta}_{M,\gamma})} \right).
\end{align*}
\hfill\ensuremath{\square}

\section{Appendix F: Proof of Theorem~\ref{thm:struct_params_pomdp}}
\label{app:struct_params_pomdp}

Let us begin with the result on the action variation in Thm.~\ref{thm:struct_params_pomdp}. We can decompose Eq.~\ref{eq:action_variation_pomdp} and find the underlying structural parameter on the true state space. We start this by realizing that $\mathbb{P}(\sigma'|\sigma,a)$ is an expected value over possible histories distributed under an arbitrary distribution over histories $\mathcal{D}_H$. Then we decompose the transition probability into its underlying components using the belief state, as done by \citet{francois-lavet_overfitting_2019}:
\begin{align*}
        \delta_{M}^{\phi} &= \max_{\sigma \in \phi(\mathcal{H})} \max_{a,a' \in \mathcal{A}} \sum_{\sigma' \in \phi(\mathcal{H})} \left| \mathbb{P}(\sigma'|\sigma,a) - \mathbb{P}(\sigma'|\sigma,a') \right| \\
        &= \max_{\sigma \in \phi(\mathcal{H})} \max_{a,a' \in \mathcal{A}} \sum_{\sigma' \in \phi(\mathcal{H})} \left| \underset{\substack{H' \sim \mathcal{D}_H: \\ \phi(H') = \sigma}}{E} \mathbb{P}(\sigma'|H',a) - \underset{\substack{H' \sim \mathcal{D}_H: \\ \phi(H') = \sigma}}{E} \mathbb{P}(\sigma'|H',a')\right| \\
    &= \max_{\sigma \in \phi(\mathcal{H})} \max_{a,a' \in \mathcal{A}} \sum_{\sigma' \in \phi(\mathcal{H})} \left| \underset{\substack{H' \sim \mathcal{D}_H: \\ \phi(H') = \sigma}}{E} \mathbb{P}(\sigma'|H',a) - \mathbb{P}(\sigma'|H',a')\right| \\
    &= \max_{\sigma \in \phi(\mathcal{H})} \max_{a,a' \in \mathcal{A}} \sum_{\sigma' \in \phi(\mathcal{H})} \left| \underset{\substack{H' \sim \mathcal{D}_H: \\ \phi(H') = \sigma}}{E} \sum_{s \in \mathcal{S}} \sum_{s' \in \mathcal{S}} b(s|H')p(s'|s,a)p(\sigma'|s',H') \right. \\ & \left. \quad -  \sum_{s \in \mathcal{S}} \sum_{s' \in \mathcal{S}} b(s|H')p(s'|s,a')p(\sigma'|s',H')\right| \\
    &= \max_{\sigma \in \phi(\mathcal{H})} \max_{a,a' \in \mathcal{A}} \sum_{\sigma' \in \phi(\mathcal{H})} \left| \underset{\substack{H' \sim \mathcal{D}_H: \\ \phi(H') = \sigma}}{E} \sum_{s \in \mathcal{S}} \sum_{s' \in \mathcal{S}}\left(p(s'|s,a) - p(s'|s,a')\right)b(s|H')p(\sigma'|s',H') \right| \\
    &\leq \max_{\sigma \in \phi(\mathcal{H})} \max_{a,a' \in \mathcal{A}} \sum_{\sigma' \in \phi(\mathcal{H})}  \underset{\substack{H' \sim \mathcal{D}_H: \\ \phi(H') = \sigma}}{E} \sum_{s \in \mathcal{S}} \sum_{s' \in \mathcal{S}}\left|p(s'|s,a) - p(s'|s,a')\right|b(s|H')p(\sigma'|s',H') \\
    &\leq \delta_{M}^S \max_{\sigma \in \phi(\mathcal{H})}\sum_{\sigma' \in \phi(\mathcal{H})} \underset{\substack{H' \sim \mathcal{D}_H: \\ \phi(H') = \sigma}}{E} p(\sigma'|s',H') \\
    &= \delta_{M}^S.
\end{align*}
We obtain the first inequality by using the triangle inequality and on the second, we use Holder's inequality on the dot product for $s \in \mathcal{S}$ to retrieve $\delta_{M}^S$. On this last inequality, we also interchange the order of summations, summing all probabilities on the support of $\sigma'$ and then taking the expectation of the constant $1$, which gives the result.

We take a similar approach to obtain the result on the value-function variation in Thm.~\ref{thm:struct_params_pomdp}. We can decompose Eq.~\ref{eq:value_function_variation_pomdp} to retrieve the equivalent parameter on the true state space:
\begin{align*}
        \kappa_{M,\gamma}^{\phi} &= \max_{\sigma,\sigma' \in \phi(\mathcal{H})} \left| V^{\pi^*_{M,\gamma}}_{M,\gamma}(\sigma) - V^{\pi^*_{M,\gamma}}_{M,\gamma}(\sigma')  \right| \\ 
         &= \max_{\sigma,\sigma' \in \phi(\mathcal{H})} \left| \sum_{s \in \mathcal{S}} b(s | \sigma) V^{\pi_\phi}_{M,\gamma}(s)  - \sum_{s \in \mathcal{S}} b(s | \sigma') V^{\pi_\phi}_{M,\gamma}(s)   \right| \\
         &\leq  \max_{\sigma,\sigma' \in \phi(\mathcal{H})} \frac{\left\|  b( \cdot | \sigma) -  b( \cdot | \sigma') \right\|_1}{2} \max_{s,s' \in \mathcal{S}} \left| V_{M,\gamma}^{\pi_\phi}(s) - V_{M,\gamma}^{\pi_\phi}(s') \right| \\ 
         &\leq  \max_{\sigma,\sigma' \in \phi(\mathcal{H})} \frac{\left\|  b( \cdot | \sigma) -  b( \cdot | \sigma') \right\|_1}{2} \left( \kappa_{M,\gamma}^{\mathcal{S}}  + \Delta_{\epsilon}^{\phi} \right)
\end{align*}
We obtain the first inequality by using lemma \ref{lemma2}. The last one is obtained by observing that $ V_{M, \gamma}^{\pi^*_{M,\gamma}}(s) \geq V_{M, \gamma}^{\pi_\phi}(s), \forall s \in \mathcal{S}$ and that by the assumption of Thm.~\ref{thm:struct_params_pomdp} we have that $- V_{M, \gamma}^{\pi_\phi}(s) \leq \Delta_{\epsilon}^{\phi} - V_{M, \gamma}^{\pi^*_{M,\gamma}}(s), \forall s \in \mathcal{S}$. \hfill\ensuremath{\square}

\end{document}